# Incorporating Dynamicity of Transportation Network with Multi-Weight Traffic Graph Convolutional Network for Traffic Forecasting

Yuyol Shin, and Yoonjin Yoon., *Member, IEEE*

*Abstract*— Traffic forecasting problem remains a challenging task in the intelligent transportation system due to its spatio-temporal complexity. Although temporal dependency has been well studied and discussed, spatial dependency is relatively less explored due to its large variations, especially in the urban environment. In this study, a novel graph convolutional network model, Multi-Weight Traffic Graph Convolutional (MW-TGC) network, is proposed and applied to two urban networks with contrasting geometric constraints. The model conducts graph convolution operations on speed data with multi-weighted adjacency matrices to combine the features, including speed limit, distance, and angle. The spatially isolated dimension reduction operation is conducted on the combined features to learn the dependencies among the features and reduce the size of the output to a computationally feasible level. The output of multi-weight graph convolution is applied to the sequence-to-sequence model with Long Short-Term Memory units to learn temporal dependencies. When applied to two urban sites, *urban-core* and *urban-mix*, MW-TGC network not only outperformed the comparative models in both sites but also reduced variance in the heterogeneous *urban-mix* network. We conclude that MW-TGC network can provide a robust traffic forecasting performance across the variations in spatial complexity, which can be a strong advantage in urban traffic forecasting.

*Index Terms*—deep learning, graph convolutional network, spatio-temporal data, traffic forecasting, transportation network

## I. INTRODUCTION

TRAFFIC forecasting is a research area in transportation engineering that has flourished over the last couple of decades and started to garner broader research interest as a key technical enabler of adaptive traffic management [1]. In the earlier efforts, the study sites for most of the studies were confined to expressways. Although the classical statistical methods such as linear regression [2], vector-autoregressive (VAR) model [3], and variations of autoregressive integrated moving average (ARIMA) model [4-5] have produced promising results, such approaches are limited in addressing complex constraints in urban areas such as signalization [6]. In complex urban areas, data-centric machine learning approaches such as the Kalman filter [7], support vector regression [8], and neural networks [9-10] have become sound alternatives. More recently, traffic volume prediction studies based on video source dataset explored various machine learning techniques such as Autoregressive Fractionally Integrated Moving Average (ARFIMA) [5] and Finite Mixture Model (FMM) [11] in an urban environment.

The recent development in deep learning techniques enabled to model the non-linearity of spatio-temporal traffic data with a vast amount of data, improving the scale and performance of the forecasting significantly. Following the earlier efforts that implemented stacked auto-encoders [12] and deep belief networks [13], many studies in the field of traffic forecasting research have explored variants of deep learning models. The convolution layers [14-19] and recurrent layers [20-25] were the major techniques that have been applied to model the time-dependent nature of traffic data. While the convolution layer [14], [17-19], [23], [26] has been the primary means of modeling spatial dependencies, the recent surge in the theory and experiments on Graph Convolutional Networks (GCN) addressed an important element in the traffic forecasting research - the transportation system is inherently embedded with graph structure. Several works proposed GCN variants with notable performance improvement [15], [24-25], [27-31].

However, the majority of existing studies adhere to a single measure, such as distance, to represent weight in the adjacency matrix. The transportation network possesses various geometric characteristics such as distance, speed limit, heading direction, and the number of lanes, which may affect the flow, speed, and density of a specific road segment and its neighbors. Resorting to a single weight lacks the ability to properly model the transportation network with multiple constraints. To adequately address the dynamicity in a transportation network and spatial non-linearity, it is necessary to build a model that can embody the multiple weighted graphs in a single architecture. In addition, the existing studies applied the same model to selected study sites, whereas the geospatial configuration of study sites could affect the model performance.

In this study, we propose Multi-Weight Traffic Graph Convolutional (MW-TGC) network and conducted experiments on speed datasets of two different urban sites. MW-TGC network was constructed to process traffic graph convolution operation [24] with multiple weights within a

Manuscript received November 14, 2019; revised May, 12, 2020 and September 11, 2020; accepted October 6, 2020. This work was supported by the National Research Foundation of Korea (NRF) grant funded by the Korea government (MSIT) (No.2017R1E1A1A01076315). *(Corresponding author: Yoonjin Yoon)*

The authors are with the Department of Civil and Environmental Engineering, Korea Advanced Institute of Science and Technology (KAIST), Daejeon 34141, South Korea (e-mail: yuyol.shin@kaist.ac.kr / yoonjin@kaist.ac.kr).

feasible amount of time and to adaptively model the spatio-temporal heterogeneity of traffic road segments. Two study sites, *urban-core* and *urban-mix*, were carefully selected to represent contrasting geospatial constraints. *Urban-core* is populated with road segments of similar road characteristics, whereas *urban-mix* consists of segments of heterogeneous weights. Ablation studies were conducted for each site to evaluate the effect of weight combinations. The main contribution of our paper is threefold:

- We proposed a genuine GCN model, MW-TGC network, which adaptively incorporates multiple inherent features of transportation networks to effectively capture the temporal and spatial variations of graph elements in urban transportation networks. Specifically, we constructed weighted adjacency matrices using speed limit, distance, and angle to conduct experiments.
- The combinations of available weights for each of the study sites with two distinctive spatial constraints were adaptively selected to achieve the best performance through ablation studies. The results indicate that the complexity of the forecasting model should also be increased as the spatial complexity increases.
- When applied to two study sites, MW-TGC network not only outperformed the comparative models but also reduced variance in the study site with heterogeneous road characteristics, showing robust performance in a larger and more complex environment.

This paper is organized as flows. In Section II, a literature review is presented, focusing on the recent development of deep learning approaches in traffic forecasting. Section III and IV explain the problem and the proposed model. In Section V, we describe the dataset and experimental setup to present the results. In Section VI, the conclusion of the study is presented as well as the scope of our future work.

## II. Literature Review

In this section, traffic forecasting studies based on deep learning models are reviewed. In Table I, studies are summarized according to their data, study area, target feature, prediction horizon, and model choice. Note that Vlahogianni *et al.* [6] presents an extensive review of traffic forecasting with traditional approaches, including earlier neural networks.

### A. Deep Learning-based Traffic Forecasting

Over the last half a decade, deep learning-based models have been actively applied to the traffic forecasting problem to model non-linear spatio-temporal dependencies in traffic data. Following the earliest works based on stacked auto-encoders [12] and deep belief networks [13], researchers have attempted to build a variety of deep learning models that could appropriately capture the spatio-temporal dependencies of traffic data.

Since traffic measures are represented as time-series data, Recurrent Neural Networks (RNN) and its variants such as Long Short-Term Memory (LSTM) [32] and Gated Recurrent Unit (GRU) [33] have naturally gained popularity to model the non-linear temporal dependencies. The vanilla LSTM has demonstrated the ability to improve the performance of traffic forecasting compared to traditional methods such as ARIMA, SVM, and Kalman filter [21-22]. When combined with Restricted Boltzmann Machine (RBM), RNN was able to find the patterns for network congestion evolution [34].

In addition to RNN, Convolutional Neural Network (CNN) was another frequently adopted technique. CNN has been utilized to extract not only spatial dependencies but also temporal dependencies of the transportation network. Ma *et al.* [14] treated the speed data matrix as images and applied CNN to capture both spatial and temporal dependencies. Through the development of time-series processing techniques and manipulation of the convolution operation [16-19], several studies have elaborated on modeling temporal dependencies through convolution layers. Wang *et al.* [19] proposed a model that captures both spatial and temporal dependencies through CNN and produced prediction using a decoder, which consists of an error-feedback recurrent layer.

Various hybrid models have been proposed to incorporate more data and model more spatio-temporal dynamics. Studies such as [20] and [26] have leveraged CNN models to model spatial dependencies and LSTM to model temporal dependencies and implemented a fusion layer to incorporate extracted features from both space and time-domain. Several studies have adopted the concept of attention mechanism by utilizing data from the past days and weeks to model the longer-term dependencies of traffic data [17], [20], [35]. Yao *et al.* [23] proposed a model that learns semantic information of the study site. For a more detailed description of recent deep learning-based traffic forecasting models, we refer the readers to [36].

### B. Graph Convolutional Network

To appropriately handle the spatial characteristics of graph-structured data with deep learning, Graph Convolutional Network (GCN) was proposed [37-39] and has been widely used in many applications such as image classification and citation network analysis.

In traffic forecasting, Li *et al.* [25] and Zhao *et al.* [29] demonstrated that a graph convolution module connected to the GRU network enhanced the ability of models to capture spatial dependencies in transportation networks. Yu *et al.* [15] and Diao *et al.* [28] used temporal convolution to extract temporal features and graph convolution to extract spatial features. Replacing recurrent units with convolution layers, the model achieved notable improvement from the perspective of computation time. Cui *et al.* [24] suggested a new type of graph convolution operation for traffic forecasting scenarios. Instead of using a Laplacian matrix, the study employed a simple adjacency matrix with sparse weight parameters. Zhang *et al.* [30] implemented an attention mechanism on traffic forecasting models leveraging traffic graph convolutional networks [24] to capture more complex non-stationary temporal dynamics. There also has been an effort to combine the Generative Adversarial Network (GAN) with the GCN to address the missing data problem in traffic data forecasting [40].

Including Lee and Rhee [27] in which the authors suggested a new graph convolution module with multiple types of weight,

TABLE I
LIST OF REFERENCES ON TRAFFIC FORECASTING STUDIES USING DEEP MODELS

| | Data | | Area | | Targeted Traffic | Prediction horizon (min)* | Model | | Additional information | |
|---|---|---|---|---|---|---|---|---|---|---|
| | Source | Resolution (min) | Site | Scale | | | Base model | Long-term information | Static 1) | Dynamic 2) |
| Lv et al., 2015 [12] | Detectors | 5 | Express | | Speed | 60 | SAE | | | |
| Huang et al., 2014 [13] | Detectors | 15 | Express | 50 | Flow | 240 | DBN | | | |
| | | | Express | 8 | | | | | | |
| Ma et al., 2017 [14] | GPS | 2 | Express | 236 | Speed | 40 | CNN | | | |
| | | | Urban | 352 | | | | | | |
| Yu et al., 2018 [15] | Detectors | 5 | Express | 12 | Speed | 45 | GCN+CNN | | • | |
| | | | Express | 1,026 | | | | | | |
| Chen et al., 2018 [16] | Video | 5 | Urban | 614 | Congestion level | 5 | CNN | • | | |
| Liu et al., 2018 [17] | Detectors | 15 | Express | 12 | Speed | 30 | CNN | | | • |
| Guo et al., 2019 [18] | Detectors | 6 | Express | 2,226 | Traffic condition (1~4)** | 60 | CNN | • | | |
| | GPS | 30 | Urban | 1,024 | Taxi flow | | | | | |
| | | 60 | Urban | 128 | Bike flow | | | | | |
| Wang et al., 2017 [19] | GPS | 5 | Express | 80 | speed | 50 | CNN | | | |
| | | | Express | 122 | | | | | | |
| Wu et al., 2018 [20] | Detectors | 5 | Express | 33 | Flow | 45 | CNN+ GRU | • | | |
| Ma et al., 2015 [21] | Detectors | 2 | Express | 2 | Speed | 2 | LSTM | | | • |
| Zhao et al., 2017 [22] | Detectors | 5 | Express | 500 | Flow | 60 | LSTM | | | |
| Yao et al., 2018 [23] | Online taxi request | 30 | Urban | 400 | Taxi demand | 30 | CNN+LSTM | | • | • |
| Cui et al., 2019 [24] | Detectors | 5 | Express | 323 | Speed | 50 | TGCN+ LSTM | | | |
| | GPS | 5 | Urban | 1,014 | | | | | | |
| Li et al., 2018 [25] | Detectors | 5 | Express | 207 | Speed | 60 | GCN + GRU | | • | |
| | | | Express | 325 | | | | | | |
| Lv et al., 2018 [26] | GPS | 5~30 | Urban | > 10,000 | Speed | 120 | CNN+LSTM | • | • | • |
| | | | Urban | ~ 1,500 | | | | | | |
| Lee and Rhee, 2019 [27] | GPS | 5 | Urban | 480 | Speed | 60 | GCN+CNN | | • | |
| | | | Urban | 455 | | | | | | |
| Diao et al., 2019 [28] | Detectors | 5 | Urban | 50 | Speed | 45 | GCN+CNN | | • | |
| | | | Express | 228 | | | | | | |
| Zhao et al., 2019 [29] | GPS | 15 | Urban | 156 | Speed | 60 | GCN+GRU | | | |
| | Detectors | 5 | Express | 207 | | | | | | |
| Zhang et al., 2019 [30] | GPS | 5 | Express | 163 | Speed | 30 | TGCN+GRU | | | • |
| | | | Urban | 242 | | | | | | |
| Liao et al., 2018 [31] | GPS | 15 | Urban | 15,073 | Speed | 120 | GCN+LSTM | • | • | • |
| Ma et al., 2015 [34] | GPS | 5/10/ 30/60 | Urban | 515 | Speed (categorized) | 60 | RBN+RNN | | | |
| Yao et al., 2019 [35] | GPS | 30 | Urban | 200 | Taxi volume | 30 | CNN+LSTM | | • | • |
| | | | Urban | 200 | Bike volume | | | | | |
| Yu et al., 2019 [40] | GPS | 5 | Urban | 37,034 | Speed | Real time | GAN+GCN | | | |
| Pan et al., 2019 [41] | GPS | 60 | Urban | 1,024 | Taxi flow | 180 | GAT+GRU | | • | • |
| | Detectors | | Express | 207 | Speed | 60 | | | | |

(*: longest prediction horizon / **: 1-smooth ~ 4-most crowded / SAE: Stacked Auto-Encoder / DBN: Deep Belief Network / TGCN: Traffic Graph Convolution Network / GAN: Generative Adversarial Network / GAT: Graph Attention)
1) Studies that used static additional data such as road structural information, and point-of-interest.
2) Studies that used dynamic additional data such as wind, precipitation, time-of-day, and holidays

there have been efforts to incorporate structural auxiliary information of road networks on graph convolutional networks [31], [41]. Although several studies tried to impose various weights on adjacency matrices using structural features of transportation networks, most of them utilized only one or a fixed number of weights, even though the significance of each feature may vary among transportation networks. Thus, one needs to consider different combinations of weights for different study sites, since the geospatial configurations of transportation networks can vary greatly even in a single urban environment.

## III. PRELIMINARIES

### A. Transportation Network Graph

In our framework, the transportation network is defined as a directed graph structure $G = (V, E, \mathbf{A})$, where $V$ is a finite set of $|V| = N$ road segments (nodes), and $E$ is a finite set of edges representing the connection between the road segments. $\mathbf{A} = \{\mathbf{A}^I, \mathbf{A}^O\}$ $(\mathbf{A}^I, \mathbf{A}^O \in \mathbb{R}^{N \times N})$ is a set of inflow and outflow adjacency matrices of graph $G$. If there exists flow from segment $i$ to $j$, then $a_{ij}^O = 1, a_{ji}^O = 0, a_{ij}^I = 0$, and $a_{ji}^I = 1$. The sum of the inflow and outflow matrix $\mathbf{A}^I + \mathbf{A}^O$ would be the adjacency matrix of the undirected graph. In the rest of the paper, we use symbol $\mathbf{A}$ for both $\mathbf{A}^I$ and $\mathbf{A}^O$, since constructions of $k$-rank adjacency matrices and weighted adjacency matrices do not differ for inflow and outflow matrices. Flow information is represented by edges $e_l$ $(l = 1, ..., L)$, where $L$ represents the total number of edges in the graph. Road segment $n$ of a transportation network $G$ produces signals $x_t^n \in \mathbb{R}^d$ $(n = 1, ..., N)$, at time $t$ where $d$ is the number of features for the signals. Fig. 1 illustrates the graph generation and adjacency matrix construction of a hypothetical road network. Note that U-turn connected road segments do not have an edge between them.

### B. Traffic Speed Forecasting

Suppose traffic speed data is generated as graph signals on transportation network $G$. At each time step $t$, let $X_t = \{x_t^1, ..., x_t^N\} \in \mathbb{R}^{N \times 1}$ be the speed profile of N road segments at time $t$. Given time $t$ and graph $G$, the traffic forecasting problem is defined as finding a model that predicts the traffic speed using historical speed, and by minimizing the error between the predicted speed and actual speed in the training dataset.

$$H: [X_{t-h+1}, ..., X_t] \rightarrow \left[ \hat{X}_{t+1}, ..., \hat{X}_{t+T_p} \right], \quad (1)$$

where $h$ is the number of time steps to be used as inputs of the model, $\hat{X}_t$ is the predicted speed profile of N road segments at time $t$, and $T_p$ is the length of the prediction horizon.

### C. Traffic Graph Convolution (TGC)

The first step of MW-TGC network is to conduct TGC operations on multiple generated weights. TGC [24] utilizes the $k$-rank ($k$-th order) adjacency matrix $\mathbf{A}^k$, in which $a_{ij}^k = r \in \mathbb{N}$ if road segment $i$ and $j$ are connected by $k$ edges by $r$ paths. The $k$-rank adjacency matrix can be simply found by $k$-th product of 1-rank adjacency matrix, $\mathbf{A}^k = \prod_{i=1}^{k} \mathbf{A}$.

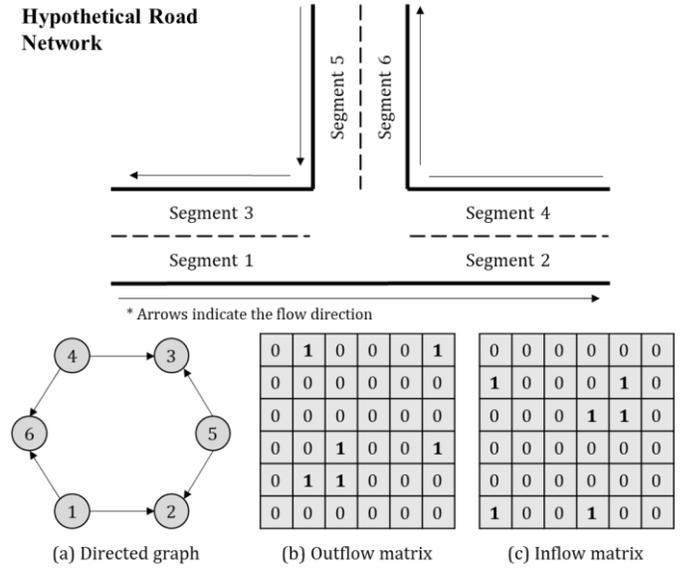

Fig. 1. Graph generation from transportation network

One of the advantages of the TGC compared to the spectral graph convolution (SGC) and its approximated variations [37-39] lies in that it alleviates the decision process for model parameters. Since the TGC only requires one N × N weight parameter matrix for each weighted adjacency matrix, the implementation of the TGC does not require a predetermined number of kernels as in the SGC. Even with the simplified decision process, the TGC can achieve a similar or better performance compared to the SGC [24].

## IV. MULTI-WEIGHT TRAFFIC GRAPH CONVOLUTIONAL (MW-TGC) NETWORK

### A. Weight Matrix Generation

A transportation network graph, just like many other graphs, can have various types of weight for consideration. In this study, we propose a framework that induces weight on transportation network graphs using the following structural features, which are incorporated adaptively in the weighted adjacency matrices.

#### 1) Distance

Distance is one of the most common weights when inducing weight on the adjacency matrix of a transportation network graph. Given the distance $d_{ij}$ between road segments $i$ and $j$ defined as the linear distance between the midpoints of the two road segments, the *distance* weighted adjacency matrix $\mathbf{W}^{dist}$ is formulated as,

$$w_{ij}^{dist} = \exp\left(-d_{ij}^2/\sigma^2\right) \text{ if } a_{ij} \neq 0. \quad (2)$$

The metric is in meters, and $\sigma^2$ is a threshold to control the distribution of the distance weight. We used 1,000 for the value of $\sigma$, transforming the metric into kilometers.

#### 2) Speed Limit

The speed limit is a key characteristic of a road segment with a strong association with speed variations. In our study site of Seoul, South Korea, the speed limit of 80km/h, 60km/h, and 30km/h represents city expressway, urban arterial roads, and local/residential roads, respectively. Since there exist wider lanes and no traffic signal in expressways, free-flow speed may differ even more than the actual speed limits among the different types of roads. In our model, we implemented three types of weights using the speed limit.

- *Speed limit-ratio* is the ratio between two consecutive segments in moving direction. It has the effect of allocating larger weight if moving from a lower speed limit segment to a higher one, and vice versa. The speed limit ratio weighted matrix $\mathbf{W}^{\mathrm{slr}}$ is defined as

$$w_{ij}^{\mathrm{slr}} = \frac{speed\ limit_j}{speed\ limit_i}, \tag{3}$$

where $speed\ limit_i$ is the speed limit of road segment $i$.

- *Speed limit-category* assigns larger weight to the road segments with a higher speed limit. The speed limit category weighted matrix $\mathbf{W}^{\mathrm{slc}}$ is defined as

$$w_{ij}^{\mathrm{slc}} = speed\ limit_j. \tag{4}$$

- *Speed limit-change* is a binary weight to flag the edges where the speed limit changes. The speed limit varying point weighted matrix $\mathbf{W}^{\mathrm{slch}}$ is defined as

$$w_{ij}^{\mathrm{slch}} = 1\ iff\ speed\ limit_i \neq speed\ limit_j. \tag{5}$$

*3) Angle*

Angle is used to represent the geometric layouts between two road segments. To assign angles, we first find the inner angle $\theta_0^{ij}$. For the road segments that are connected by ranks higher than 1, we extend the endpoint of the former road segment and the starting point of the latter one and calculated the angle between the extended lines. After $\theta_0^{ij}$ is defined, the angle weighted matrix $\mathbf{W}^{\mathrm{angle}}$ is defined as

$$w_{ij}^{\mathrm{angle}} = \exp\left(-1/\left|\pi - \theta_0^{ij}\right|\right). \tag{6}$$

*4) Plain*

In plain adjacency matrix $\mathbf{A}^k = (a_{ij}^k)$, $a_{ij}^k$ indicates the number of alternative paths from road segment $i$ to $j$.

When all the weights and directions are counted, 6 (number of weights) × 2 (number of flow direction) weighted adjacency matrices for each rank can be generated.

### B. Traffic Graph Convolution with Multiple Weights

The model structure of MW-TGC network is shown in Fig. 2. First, we define $k$-rank weighted adjacency matrix for TGC operation as

$$\widetilde{\mathbf{W}}^k = \mathrm{Clip}(\mathbf{W}^k + \mathbf{I}), \tag{7}$$

where $\mathrm{Clip}(\cdot)$ is a function to confine values of elements of $\widetilde{\mathbf{W}}^k$ between 0 and 1. The clipping procedure is done to adjust the various weighted adjacency matrices within the same scale.

The graph convolution at time $t$ with $k$-rank weighted adjacency matrix $\widetilde{\mathbf{W}}^k$ is defined as

$$\mathrm{GC}_t^k = X_t *_g \widetilde{\mathbf{W}}^k = \left(\mathbf{W}_{gc^k} \odot \widetilde{\mathbf{W}}^k\right) X_t, \tag{8}$$

where $\odot$ is the Hadamard product, or the element-wise matrix multiplication, $\mathbf{W}_{gc^k} \in \mathbb{R}^{N \times N}$ is the learnable weight parameter matrix for the $k$-rank adjacency matrix $\widetilde{\mathbf{W}}^k$, and $X_t \in \mathbb{R}^N$ denotes the input traffic speed at time $t$. In Fig. 2(a), the implementation of MW-TGC operation is shown. When $c$ types of weighted adjacency matrices and up to $k$-rank adjacency are considered, the output of MW-TGC operation on graph signal at time $t$ would be

$$\mathbf{GC}_t^{\{k\}} = \left[\mathrm{GC}_t^{1,1}, ..., \mathrm{GC}_t^{c,1}, ..., \mathrm{GC}_t^{1,k}, ..., \mathrm{GC}_t^{c,k}\right] \in \mathbb{R}^{N \times ck}, \tag{9}$$

where $c$ is the number of considered weights at each rank multiplied by 2. After MW-TGC operation, Rectified Linear Unit (ReLU) activation function is applied to the output, $\mathbf{GC}_t^{\{k\}}$.

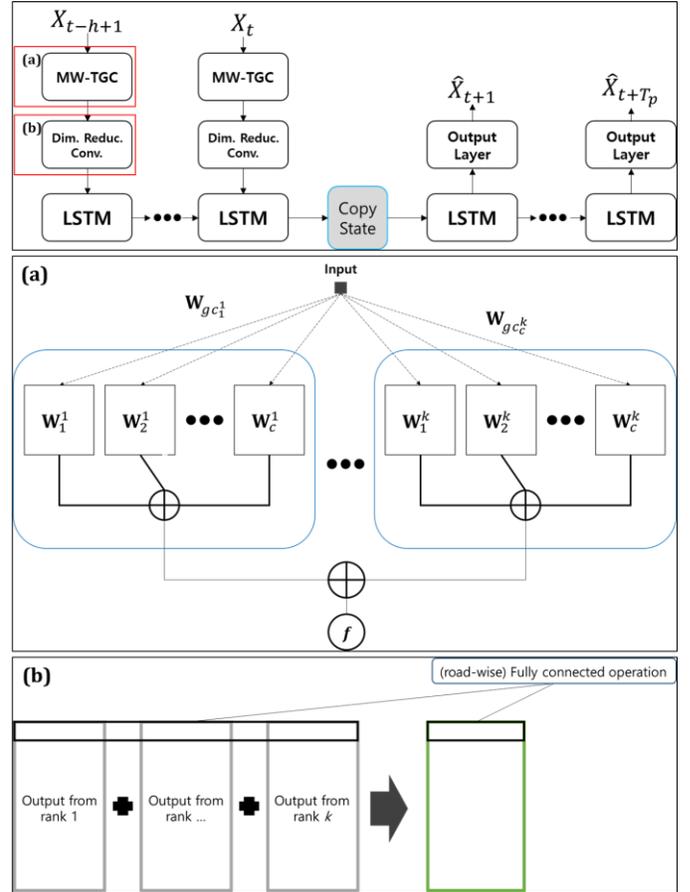

Fig. 2. Architecture of MW-TGC network. (a) Graph convolution operation for MW-TGC; (b) Dimension reduction convolution operation

To incorporate the multiple weighted adjacency matrices, the Dimension Reduction Convolution (DRC) is conducted to reduce the size of the raw MW-TGC operation output from 2(a) to a computationally feasible size (Fig. 2(b)). The DRC operation is defined as

$$h_{dr}\left(\mathbf{GC}_t^{\{k\}}\right) = \mathbf{GC}_{dr_t}^{\{k\}} = \mathbf{GC}_t^{\{k\}} *_{conv} \Gamma \in \mathbb{R}^{N \times C_{out}}, \tag{10}$$

In essence, convolution operation with kernel $\Gamma \in \mathbb{R}^{1 \times ck \times C_{out}}$ is carried out on each row $\left(\mathbf{GC}_t^{\{k\}}\right)_{i:} \in \mathbb{R}^{ck}, (i = 1, ..., N)$ to preserve the spatially extracted information of road segment $i$ from its adjacent segments. Here, $C_{out}$ is the number of output features.

In addition, ablation is applied to experiment the weight adaptation to various weight complexity. The technique has been widely adopted to justify the model architecture [23], [31], [41].

### C. Spatio-Temporal Seq2Seq Model Using LSTM

To learn the dynamicity of temporal dependencies, spatial dependencies are modeled with MW-TGC module at each time step $t$, and the output sequence is then fed to the spatio-temporal sequence-to-sequence (Seq2Seq) model [42] implemented using LSTM [32]. Hochreiter [32] proposed LSTM to address vanishing gradient problem of RNN. By adopting tanh activation function, an LSTM unit can keep the value of gradients larger than an RNN unit with sigmoid activation function can. In addition, the cell state implemented in LSTM helps recurrent layers to pass long-term memory through long sequential learning.

In the encoder units of the Seq2Seq model, the gates are structured as the input vector replaced by the reduced MW-TGC outputs at time step $t$, $\mathbf{GC}_{dr_t}^{\{k\}} \in \mathbb{R}^{C_{out}N}$, where $\mathbf{GC}_{dr_t}^{\{k\}}$ is reshaped into a vector. The input gate $i_t$, forget gate $f_t$, output gate $o_t$, and candidate cell state $\tilde{C}_t$ at time step $t$ are defined as follows:

$$i_t = \sigma\left(\mathbf{GC}_{dr_t}^{\{k\}}\mathbf{W}_{i_{GC}} + h_{t-1}\mathbf{W}_{i_{hidden}} + b_i\right) \quad (11)$$

$$f_t = \sigma\left(\mathbf{GC}_{dr_t}^{\{k\}}\mathbf{W}_{f_{GC}} + h_{t-1}\mathbf{W}_{f_{hidden}} + b_f\right) \quad (12)$$

$$o_t = \sigma\left(\mathbf{GC}_{dr_t}^{\{k\}}\mathbf{W}_{o_{GC}} + h_{t-1}\mathbf{W}_{o_{hidden}} + b_o\right) \quad (13)$$

$$\tilde{C}_t = \tanh\left(\mathbf{GC}_{dr_t}^{\{k\}}\mathbf{W}_{C_{GC}} + h_{t-1}\mathbf{W}_{C_{hidden}} + b_C\right), \quad (14)$$

where $\mathbf{W}_{i_{GC}}, \mathbf{W}_{f_{GC}}, \mathbf{W}_{o_{GC}}$ and $\mathbf{W}_{C_{GC}} \in \mathbb{R}^{(C_{out}N) \times h_{size}}$ are the weight parameter matrices for the mapping from the inputs to the hidden states for the corresponding gates of LSTM and the candidate cell state; $h_t \in \mathbb{R}^{h_{size}}$ is the hidden state with $h_{size}$ neurons; $\mathbf{W}_{i_{hidden}}, \mathbf{W}_{f_{hidden}}, \mathbf{W}_{o_{hidden}}$ and $\mathbf{W}_{C_{hidden}} \in \mathbb{R}^{h_{size} \times h_{size}}$ are the weight parameter matrices for mapping from previous hidden state to the next hidden state in the corresponding gates; and $b_i, b_f, b_o$ and $b_C \in \mathbb{R}^{h_{size}}$ are the bias vectors. Now, the cell state and hidden state at time step $t$ is defined as

$$C_t = f_t \odot C_{t-1} + i_t \odot \tilde{C}_t \quad (15)$$

$$h_t = o_t \odot \tanh(C_t). \quad (16)$$

In the decoder units, $\mathbf{GC}_{dr_t}^{\{k\}}$ is replaced by $\hat{X}_{t-1}$ that is predicted speed at time $t-1$, and the weight parameters are replaced accordingly. $\hat{X}_{t-1}$ is achieved through a output layer defined as $f(h_{t-1}^d) \rightarrow \hat{X}_{t-1}$, where $h_{t-1}^d$ is the hidden state of the decoder units with $h_{size}$ neurons.

## V. EXPERIMENT

### A. Datasets

To evaluate the performance of the proposed model, we applied the model to the average taxi speed datasets of Seoul, South Korea. In South Korea, a government agency named TOPIS aggregates the raw GPS tracking data collected from DTG (Digital Tacho Graph) on Seoul taxis into 5 min resolution, which meets the minimum update time required to ensure traveler information systems to be beneficial in congested transportation networks [43].

Our dataset contains the 5 min average speed from April 1 to April 30, 2018, and we used the first 21 days as the training set, the following 2 as the validation set, and the remaining 7 days as the test set. The citywide network has 4,774 road segments in total[1], and we extracted two sites for the experiment, namely, *urban-core* and *urban-mix*, as shown on the map in Fig. 3.

TABLE II
The Speed Limit Distribution in Urban-Mix

| Speed limit (km/h) | 40 | 50 | 60 | 70 | 80 | 100 |
|---|---|---|---|---|---|---|
| Count | 70 | 1 | 818 | 37 | 76 | 5 |

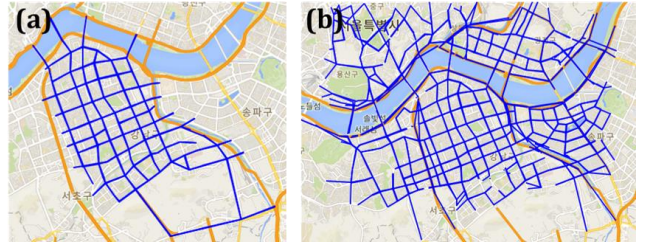

Fig. 3.  Network of *urban-core* (A) and *urban-mix* (B).

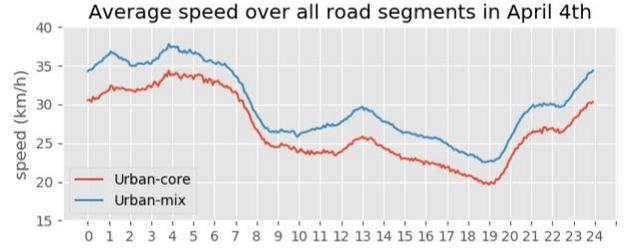

Fig. 4.  Average speed over all road segments on a typical weekday (Wednesday, April 4, 2018)

*Urban-core* is designed to include road segments with homogeneous speed limits. The site is extracted from Gangnam, Seoul, South Korea, which is one of the areas of the highest traffic in the country. It is populated with 304 road segments with an identical speed limit of 60km/h, where the mean speed is 27.63 km/h, and the standard deviation is 11.25 km/h. The resulting graph has 304 nodes and 696 edges both in the inflow and outflow graphs.

*Urban-mix* is a spatial expansion of *urban-core* to include the road segments with heterogeneous speed limits. It includes expressway connecting the east and the west end of the city, urban arterials, alleys, and bridges, spanning from Cheongdam Bridge to Dongjak Bridge. The network of *Urban-mix* contains 1,007 road segments and 2,295 edges for each of the inflow and outflow graph. The speed limit ranges from 40 km/h to 100km/h, as summarized in Table II. The mean speed is 30.69 km/h, and the standard deviation is 16.91 km/h. The network scale of 304 and 1,007 segments in both sites is relevant compared to existing GCN studies [15], [24-25], [27-31], [40].

In Fig. 4, 5 min speed of Wednesday, April 4 is shown, averaged over all road segments. One can observe that both sites have similar trends, with *urban-mix* being elevated. One of the motivations to study two sites of contrasting geospatial configuration is to obtain comparative insight of graphical convolution performance with multi-weight consideration, even when the overall trend seems nearly identical.

### B. Metric, Comparative Models and Settings

For the performance measure of our experiments, we choose Root Mean Squared Error (RMSE), Mean Absolute Deviation (MAD), Mean Absolute Percentage Error (MAPE), and Mean Absolute Scaled Error (MASE). The evaluation metrics are

---

[1] The dataset is available at TOPIS (https://topis.seoul.go.kr)

defined as the following:

$$\text{RMSE} = \sqrt{\sum_{i=1}^{N} \sum_{j=1}^{T} \frac{\left(\hat{X}_{ij} - X_{ij}\right)^2}{TN}}, \quad (17)$$

$$\text{MAPE} = \frac{1}{TN} \sum_{i=1}^{N} \sum_{j=1}^{T} \frac{\left|\hat{X}_{ij} - X_{ij}\right|}{X_{ij}}, \quad (18)$$

$$\text{MAD} = \frac{1}{TN} \sum_{i=1}^{N} \sum_{j=1}^{T} \left|\hat{X}_{ij} - X_{ij}\right| \quad (19)$$

$$\text{MASE} = \frac{1}{N} \sum_{i=1}^{N} \left( \frac{\sum_{j=1}^{T} |\hat{X}_{ij} - X_{ij}|}{\frac{1}{T-1} \sum_{j=2}^{T} |X_{ij} - X_{i,j-1}|} \right) \quad (20)$$

where $T$ is the total number of predicted time steps, N is the number of road segments, $\hat{X}_i$ is the prediction value, and $X_i$ is the actual value.

To adaptively assign the weights to each study site, we assessed the performance of the multi-weight model by repeating the experiment with various combinations of the weighted adjacency matrices. Based on the single-weight model performance, we selected the combinations of the weights as presented in Section V. D.

Also, we compared MW-TGC network with the following comparative models (2 statistical, 2 simple deep learning, and 2 graph convolution networks) for the performance evaluation:

- Historical Average (HA) depending solely on the speed of the most recent time step used in the prediction task,
- Vector Auto-Regressive model (VAR) [3] with 2 lags,
- Feed-forward Neural Network (FNN) with 2 hidden layers where the number of neurons for hidden layers is 8 * N and 4 * N (N is the number of segments in study sites),
- Sequence to Sequence Learning (Seq2Seq) [42] with 1 recurrent layer, and LSTM units as recurrent units where the number of neurons for the hidden layer is N,
- Spatio-Temporal Graph Convolutional Networks (ST-GCN) [15], and
- Traffic Graph Convolutional LSTM [24].

For the experiment of MW-TGC network, we set the number of historical time steps $h$ used for forecasting task to 12, and the length of the prediction horizon $T_p$ to 12. In other words, the model took historical 1 hour of speed data as input and predicted speed for the next 1 hour. For analysis, we measured the model performance at the forecasting horizon of 6 (30 min), 9 (45 min), and 12 (60 min) time steps. The maximum rank of the adjacency matrices $k$ was 3 for the final outcomes. The number of elements for the set of weighted adjacency matrices $\mathbb{W}$ has varied from 6 (1 weight, 2 flow directions, and 3 ranks) to 36 (6 weights, 2 flow directions, and 3 ranks). Thus, applying (3) to our framework, the graph convolution would be

$$\mathbf{GC}^{\{k\}} = X_t *_g \mathbb{W} = \text{CAT}\left( \left( \mathbf{W}_{gc^{c,k}} \odot \widetilde{\mathbf{W}}^{c,k} \right) X_t \right) \forall (c \text{ and } k), \quad (21)$$

where $\text{CAT}(\cdot)$ is the concatenation of the elements, and $c$ is the number of weighted adjacency matrices for each rank. The dimension reduction convolution $h_{dr}(\mathbf{GC}^{\{k\}})$ has parameter $C_{out} = 4$, thus a kernel size of $\Gamma \in \mathbb{R}^{1 \times c k \times C_{out}}$.

$$h_{dr}\left(\mathbf{GC}_t^{\{k\}}\right) = \mathbf{GC}_t^{\{k\}} *_{conv} \Gamma = \mathbf{GC}_{dr_t}^{\{k\}} \in \mathbb{R}^{N \times 4} \quad (22)$$

The output of this operation was fed to sequence to sequence

LSTM model. For the Seq2Seq part, the number of the hidden unit is N × 2, which is two times the number of road segments, and the number of recurrent layers is 1. The final training settings are L2 loss function, RMSprop optimizer, a learning rate of 1e-3 with learning rate decay of 0.7 at every 5 steps, and batch size of 50 for *urban-core*, and RMSprop optimizer, learning rate of 1e-3 with decay of 0.9 at every 5 steps, and batch size of 50 for *urban-mix*. We adopted early stopping method to make sure that the model converge. Using the validation set, we checked that there was no overfitting problem with these parameters. For the loss function, we used L2 loss. For all other deep learning models, the same parameters yielded the best performance. To make a fair comparison, we implemented LSTM to all recurrent models. All the deep learning models are evaluated on a single NVIDIA TITAN RTX with 24GB memory (GPU) and Intel(R) Xeon(R) CPU ES-2630 v4 @ 2.20GHz (CPU).

### C. Evaluation of MW-TGC network

#### 1) Comparative study

To compare the performance of the proposed model with the six comparative methods, we conducted experiments on the two study sites. Table III and Table IV show the performance of each model on the two study sites. For *urban-core*, the combination of *plain*, and *speed limit-ratio* weights yielded the best results. In contrast, the combination of *speed limit-ratio*, *speed limit-category*, *speed limit-change*, and *distance* weights yielded the best results for *urban-mix*.

From Table III and Table IV, MW-TGC network showed the best performance over the other comparative models in all four evaluation metrics. From the result of the experiment, we observed that even a simple deep learning model (FNN) outperformed the traditional statistical methods (HA, and VAR). Although the statistical models may have an advantage in clarity in the analysis of the derived parameters, the deep learning models indeed produced more accurate predictions. In *urban-core*, the performance of the GCN comparative model (ST-GCN) roughly matched the performance of a simple FNN model. In contrast, as the size of the road network and the heterogeneity among road segments increased in *urban-mix*, the FNN model produced better performance than ST-GCN. In the case of the proposed MW-TGC network, the model achieved a more accurate forecast than all other deep learning models in both study sites. This result indicates that the graph convolution with multiple weights consideration scheme efficiently captured the dynamicity in the transportation network graph. In addition, by comparing the performance gain in each site, we observed that the model achieved larger performance improvement in *urban-mix*. For *urban-mix*, RMSE for 1-hour prediction improved by 50.1% compared to the HA, whereas for *urban-core*, the improvement for the same task was 30.3%. MW-TGC network helped achieve more performance gain in the network with heterogeneous road characteristics than in the network with similar road characteristics. This result indicates that for a network with similar road characteristics, we may need information other than structural features implemented in this study to achieve further performance improvement.

TABLE III
FORECASTING ERROR ON URBAN-CORE (30/45/60MIN)

| | RMSE (km/h) | MAPE (%) | MAD (km/h) | MASE |
|---|---|---|---|---|
| HA | 5.159/5.538/5.832 | 13.688/15.161/16.257 | 3.468/3.790/4.030 | 1.217/1.346/1.443 |
| VAR | 4.446/4.806/5.126 | 12.109/13.369/14.448 | 3.042/3.337/3.595 | 1.076/1.193/1.295 |
| FNN | 4.030/4.061/4.108 | 11.326/11.479/11.627 | 2.754/2.786/2.821 | 0.971/0.985/0.999 |
| Seq2Seq | 4.068/4.080/4.099 | 11.391/11.433/11.511 | 2.772/2.782/2.799 | 0.977/0.981/0.988 |
| ST-GCN | 4.089/4.369/4.597 | 11.538/12.648/13.249 | 2.808/3.025/3.199 | 0.992/1.080/1.148 |
| TGC-LSTM | 4.029/4.049/4.086 | 11.295/11.355/11.496 | 2.749/2.765/2.799 | 1.009/1.016/1.032 |
| MW-TGC | 4.022/4.041/4.064 | 11.197/11.221/11.354 | 2.747/2.761/2.782 | 0.967/0.973/0.982 |

TABLE IV
FORECASTING ERROR ON URBAN-MIX (30/45/60MIN)

| | RMSE (km/h) | MAPE (%) | MAD (km/h) | MASE |
|---|---|---|---|---|
| HA | 6.031/6.664/7.130 | 14.246/15.779/16.998 | 3.758/4.134/4.431 | 1.323/1.461/1.571 |
| VAR | 5.428/5.991/6.425 | 12.643/13.921/14.997 | 3.340/3.684/3.983 | 1.176/1.302/1.412 |
| FNN | 4.237/4.352/4.631 | 10.845/11.030/11.480 | 2.761/2.834/3.016 | 0.974/1.000/1.066 |
| Seq2Seq | 4.198/4.209/4.273 | 10.699/10.731/10.858 | 2.642/2.646/2.680 | 0.987/0.986/1.000 |
| ST-GCN | 4.707/5.237/5.618 | 11.444/12.497/13.253 | 2.944/3.204/3.396 | 1.037/1.129/1.197 |
| TGC-LSTM | 3.506/3.519/3.632 | 9.171/9.233/9.525 | 2.354/2.366/2.439 | 0.868/0.873/0.904 |
| MW-TGC | 3.405/3.424/3.559 | 8.957/9.048/9.430 | 2.297/2.314/2.404 | 0.847/0.855/0.894 |

TABLE V
RESULT OF THE DIEBOLD-MARIANO TEST ON 1 HOUR FORECAST

| | *Urban-Core*<br>MW-TGC | *Urban-Mix*<br>MW-TGC |
|---|---|---|
| HA | 5.97e-3** | 4.58e-5** |
| VAR | 8.42e-3** | 1.40e-5** |
| FNN | 0.358 | 1.22e-3** |
| Seq2Seq | 0.366 | 2.23e-3** |
| ST-GCN | 4.84e-2* | 3.10e-4** |
| TGC-LSTM | 0.327 | 0.223 |

\* p-value < 0.1
\*\* p-value < 0.05

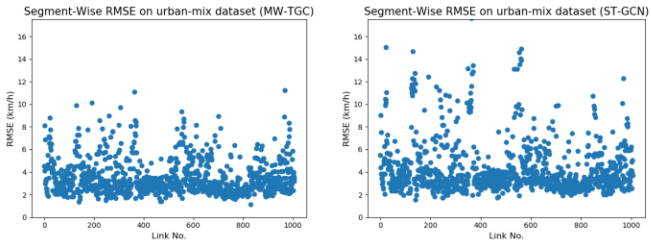

Fig. 5. RMSE on each road segment of urban-mix using the MW-TGC network (*left*) and ST-GCN (*right*). The prediction horizon for these results is 30 min.

Another implication is that the performance of the proposed model may be enhanced by having additional appropriate information about the transportation network, such as the number of lanes, the number of signals, and land uses.

In addition, we conducted the Diebold-Mariano test [44] to show that the forecasting outcome was statistically significant. The test was conducted on 1-hour prediction outcomes based on Mean Squared Error (MSE) of MW-TGC network against the comparative models. The results in Table V showed that MW-TGC network in *urban-mix* achieved consistent statistical significance compared to all the comparative models, except for MW-TGC & TGC-LSTM comparison. However, it failed to show significance against the FNN and the recurrent models in *urban-core*.

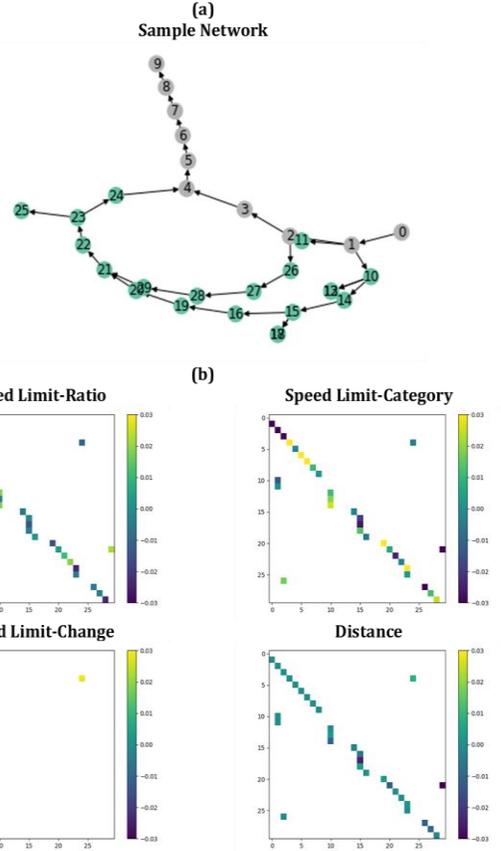

Fig. 6. (a) Sample network from urban-mix study area. The grey (lighter) nodes (1~9) indicate road segments with a speed limit of 80 km/h, and the green (darker) ones (10~29) indicate that of 60 km/h. (b) The element-wise multiplication of weighted inflow adjacency matrices and corresponding weight parameters on the sample network.. The color ranges from blue (darker) to yellow (lighter). The weight parameter matrices correspond to *speed limit-ratio*, *speed limit-category*, *speed limit-change*, and *distance* (top left to bottom right).

### 2) Localized performance

To evaluate the performance of MW-TGC network on each road segment, we made a comparison between MW-TGC network with another GCN model, ST-GCN. The plots in Fig. 5 are RMSE of 30 min forecast made on individual road segments. For ST-GCN, 68 road segments were two standard deviations off from the mean, and for MW-TGC network, 51 were. This result implies that the proposed model performed with higher precision compared to ST-GCN. Among the road segments with the 50 highest RMSE of ST-GCN, 11 segments had the speed limit of 60km/h, 1 of 70km/h, 37 of 80km/h, and 1 of 100km/h. In contrast, for MW-TGC, 22 road segments had the speed limit of 80km/h, and 23 had 60km/h. Although it was still from the ratio of 60km/h to 80km/h road segments (818 to 76), the proposed model indeed reduced variance in heterogeneous *urban-mix*, showing robust performance compared to ST-GCN.

### 3) Model Parameters

To assess the impacts of edges that connect pairs of road segments, we visualized the multiplication of each rank 1 inflow weighted adjacency matrix and corresponding weight parameter for sampled road segments (Fig. 6). In the *distance* inflow matrix, all the edges were converged to have similar weight products. In the *speed limit-change*, the single change of $60 \rightarrow 80$ km/h (node 24 $\rightarrow$ 4) and two changes of $80 \rightarrow 60$ km/h (node 1 $\rightarrow$ 10 & node 2 $\rightarrow$ 26) showed a similar magnitude but the opposite sign. The change $80 \rightarrow 60$ km/h in edge connecting node 1 to node 11 had the weight product of a smaller magnitude. The weight products for *speed limit-ratio* and *speed limit-category* weights exhibited a mixed result with cases of weight products converging opposite signs and the others converging to have similar values. Such results show that each layer of the MW-TGC operation can learn different spatial dependencies of the transportation network effectively.

### D. Weight Adaptation Using Ablation Technique

Ablation is utilized for weight adaption in two sites with different geospatial complexity. Note that the *speed limit category* weight was excluded in the *urban-core* ablation since it is identical to the *speed limit-ratio* weight with the identical speed limit on the site. Computation times for each epoch and the number of epochs needed to reach convergence of the comparative models and MW-TGC network with different numbers of ablated weights are shown in Fig. 7, and Table VI. The forecasting errors for the prediction horizon of 60 min is presented in Table VII for *urban-core*, and in Table VIII for *urban-mix*. In the ablation, three training processes were repeated for each weight combination to reduce the randomness of the outcomes.

As shown in Table VII, the performance gain of using multiple weighted adjacency matrices is not apparent in *urban-core*. The highest RMSE gain is only 0.34% between the single weight of *speed limit-ratio* and the combined weight of *plain* and *speed limit-ratio* weights. On the contrary, multiple weights enabled the model to learn heterogeneous features of *urban-mix* network more effectively, showing a clear advantage of the multi-weight graph convolution. Comparing the single weight of *speed limit-ratio* with the four-weight combination of *speed limit-ratio*, *speed limit-category*, *speed limit-change*, and

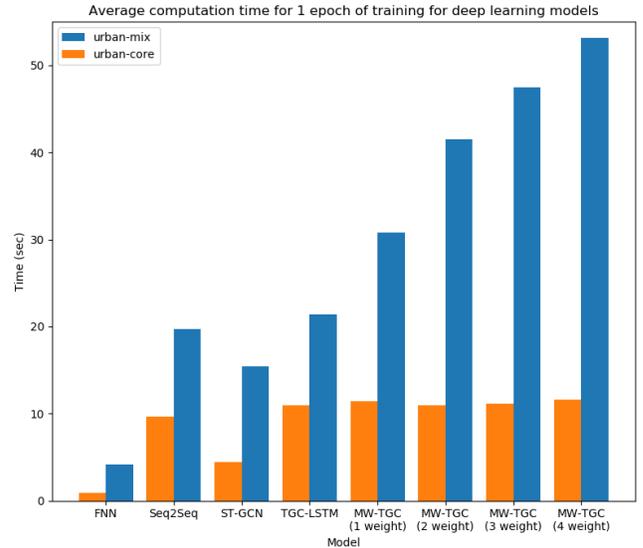

Fig. 7. Average computation time for training 1 epoch for deep learning models

TABLE VI
THE NUMBER OF EPOCHS NEEDED TO REACH CONVERGENCE
(AVERAGED OVER 3 TRIALS)

|  | Urban-Core | Urban-Mix |
|---|---|---|
| FNN | 64.3 | 75.7 |
| Seq2Seq | 66.7 | 183.0 |
| ST-GCN | 47.0 | 50.7 |
| TGC-LSTM | 39.0 | 248.0 |
| MW-TGC (best single weight) | 45.3 | 152.7 |
| MW-TGC (best) | 45.7 | 161.3 |

*distance* weights, the proposed model achieved a 4.69% gain in RMSE. The ablation also contributed to increase the forecasting precision in *urban-mix* by reducing the standard deviation of the training results. The standard deviation of the RMSE from 3 trials decreased from 0.364 to 0.032 in *urban-mix*. In *urban-core*, precision also stayed at a similar level.

The ablation results clearly show that the effect of the weight choice varies with different geospatial configuration. Whereas the *angle* weight yielded a performance comparable to *speed limit-ratio* weight in *urban-core*, using the same weight resulted in lower performance than other single-weight applications in *urban-mix*.

## VI. CONCLUSION AND FUTURE WORK

In this paper, we proposed a novel traffic forecasting model to incorporate various structural features of the transportation network to model spatio-temporal dependencies within the traffic graph convolution framework. The model enables to feed multiple types of weights, or information, into a single graph convolutional network. Experiments were carried out on two study sites with different geospatial configurations, and results demonstrated that MW-TGC network outperformed other state-of-the-art graph convolution models in both sites. The larger performance gain in heterogeneous *urban-mix* compared to homogeneous *urban-core* implies that the information on the structural property is well incorporated in the GCN framework through multi-weight implementation. Ablation study shows

TABLE VII
TEST RESULT ON 60 MIN FORECAST WITH DIFFERENT WEIGHT ABLATIONS (URBAN-CORE)

| | RMSE (km/h) | MAPE (%) | MAD (km/h) | MASE |
|---|---|---|---|---|
| Plain | 4.089 ± 0.041 | 11.617 ± 0.282 | 2.807 ± 0.040 | 1.014 ± 0.033 |
| SL-R | 4.077 ± 0.010 | 11.440 ± 0.028 | 2.795 ± 0.011 | 0.988 ± 0.004 |
| Dist | 4.124 ± 0.081 | 11.645 ± 0.301 | 2.832 ± 0.067 | 1.003 ± 0.027 |
| Angle | 4.078 ± 0.003 | 11.432 ± 0.029 | 2.793 ± 0.002 | 0.987 ± 0.001 |
| **Plain & SL-R** | 4.063 ± 0.003 | 11.388 ± 0.023 | 2.781 ± 0.004 | 0.984 ± 0.003 |
| Plain & Dist | 4.064 ± 0.005 | 11.354 ± 0.038 | 2.782 ± 0.005 | 0.982 ± 0.003 |
| Plain & SL-R & Angle | 4.068 ± 0.006 | 11.324 ± 0.102 | 2.781 ± 0.006 | 0.982 ± 0.002 |
| Plain & SL-R & Dist & Angle | 4.075 ± 0.014 | 11.394 ± 0.006 | 2.785 ± 0.007 | 1.047 ± 0.001 |

SL-R: speed limit-ratio / Dist: distance

TABLE VIII
TEST RESULT ON 60 MIN FORECAST WITH DIFFERENT WEIGHT ABLATIONS (URBAN-MIX)

| | RMSE (km/h) | MAPE (%) | MAD (km/h) | MASE |
|---|---|---|---|---|
| Plain | 4.048 ± 0.645 | 10.490 ± 1.330 | 2.666 ± 0.345 | 1.004 ± 0.153 |
| SL-R | 3.726 ± 0.364 | 9.771 ± 0.806 | 2.488 ± 0.196 | 0.925 ± 0.075 |
| SL-Cat | 4.048 ± 0.675 | 10.446 ± 1.449 | 2.662 ± 0.365 | 0.999 ± 0.165 |
| Dist | 4.208 ± 0.676 | 10.956 ± 1.436 | 2.787 ± 0.356 | 1.060 ± 0.153 |
| Angle | 4.175 ± 0.490 | 10.685 ± 1.056 | 2.729 ± 0.261 | 1.027 ± 0.118 |
| Plain & SL-R | 3.606 ± 0.059 | 9.524 ± 0.165 | 2.428 ± 0.035 | 0.902 ± 0.014 |
| SL-R & Angle | 4.031 ± 0.783 | 10.378 ± 1.545 | 2.666 ± 0.434 | 1.007 ± 0.189 |
| SL-Cat & SL-Ch & Angle | 3.603 ± 0.079 | 9.545 ± 0.210 | 2.428 ± 0.048 | 0.902 ± 0.019 |
| SL-R & SL-Cat & SL-Ch & Angle | 3.572 ± 0.078 | 9.461 ± 0.232 | 2.410 ± 0.046 | 0.895 ± 0.017 |
| **SL-R & SL-Cat & SL-Ch & Dist** | 3.559 ± 0.032 | 9.430 ± 0.099 | 2.404 ± 0.018 | 0.894 ± 0.006 |

SL-R: speed limit-ratio / SL-Cat: speed limit-category / SL-Ch: speed limit-change / Dist: distance that increasing complexity with a higher number of weights is most beneficial in the environment of higher geospatial complexity of *urban-mix*. MW-TGC not only captures the spatial heterogeneity within a transportation network but also can infer such characteristics even with the increased complexity through effective weight combinations. Our results strongly suggest that adaptively incorporating various weights is beneficial according to the geospatial configuration of study sites in applying the GCN models.

In future work, we plan to further assess the structural dynamics of the transportation network based on network science. Node centrality measures such as betweenness and closeness centrality may be more helpful than the raw structural information on enhancing the learning ability of a model. The layered complex network concept [45-46], which builds separate graphs according to different edge weights, may be incorporated into our model to better reflect the dynamics of the networks. In addition, we plan to construct a model to learn long-term periodicity by incorporating data longer in the past (i.e., speed of day before) to improve the forecasting accuracy. We plan to acquire raw GPS data and process it to traffic measurements to increase the amount and adjust the resolution of the data. Using video-sourced data and techniques to process such data [5], [11], we would be able to verify the credibility of the processed GPS data [47]. Future works may scale to broader applications in the transportation domain, such as taxi-demand prediction and travel time estimation.

REFERENCES AND FOOTNOTES

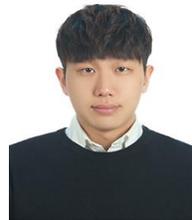

**Yuyol Shin** received the B.S. degree in civil and environmental engineering from Korea Advanced Institute of Science and Technology (KAIST), Daejeon, South Korea in 2016. He is currently pursuing the Ph.D. degree in civil and environmental engineering from KAIST, Daejeon, South Korea. His research interests include spatio-temporal data mining, artificial intelligence and transportation network analysis.

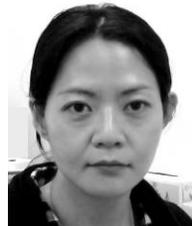

**Yoonjin Yoon** (M' 17) received the B.S. degree in mathematics from Seoul National University, Seoul, South Korea, in 1996 and dual M.S. degree in Computer Science and Management Science and Engineering from Stanford University, Stanford, CA, USA in 2000 and 2002. She received her Ph.D. degree in Civil and Environmental Engineering from University of California, Berkeley, CA, USA in 2010.

Since 2011, she has been an Assistant Professor in the department of Civil and Environmental Engineering at Korea Advanced Institute of Science and Technology (KAIST) in Daejeon, South Korea. Before arriving at KAIST, she was a Graduate Student Researcher at the National Center of Excellence in Air Transportation Operations Research (NeXTOR) at University of California, Berkeley, CA, USA from 2005 to 2010. Previous research experience includes Research Assistant in the Artificial Intelligence Center at SRI International, Menlo Park, CA, USA in 1999, and Center of Reliability Computing at Stanford University, Stanford, CA, USA from 2000 to 2002. Her main research area is the traffic management of both manned and unmanned vehicles using stochastic optimization. Her recent research efforts include data driven driving behavior analysis, and autonomous vehicle traffic flow management using large scale driving data.